\documentclass[letterpaper]{article} 
\usepackage{aaai25}  
\usepackage{times}  
\usepackage{helvet}  
\usepackage{courier}  
\usepackage[hyphens]{url}  
\usepackage{graphicx} 
\usepackage{natbib}  
\usepackage{caption} 
\usepackage{algorithm}
\usepackage{algorithmic}

\usepackage{newfloat}
\usepackage{listings}
\DeclareCaptionStyle{ruled}{labelfont=normalfont,labelsep=colon,strut=off} 
\floatstyle{ruled}
\newfloat{listing}{tb}{lst}{}
\floatname{listing}{Listing}
\nocopyright 

\usepackage{amsmath}
\usepackage{tabularx}
\usepackage{booktabs}
\usepackage{url}
\usepackage{ragged2e}
\usepackage{hyphenat}
\usepackage{microtype}
\usepackage{soul}

\newcommand\pda{$\textsc{PDA}$\ }
\newcommand\pa{$\textsc{PDA}$}

\title{Pyramid-Driven Alignment: Pyramid Principle Guided Integration of Large Language Models and Knowledge Graphs}
\author {
    Lei Sun\textsuperscript{\rm 1}, 
    Xinchen Wang,
    Youdi Li\textsuperscript{\rm 1},
}
\affiliations {
    \textsuperscript{\rm 1}Panasonic Connect, Tokyo, Japan\\
    sun.lei@jp.panasonic.com, xinchen0803@gmail.com, ri.yutei@jp.panasonic.com
}

\usepackage{bibentry}

\begin{document}

\maketitle

\begin{abstract}
Large Language Models (LLMs) possess impressive reasoning abilities but are prone to generating incorrect information, often referred to as hallucinations. While incorporating external Knowledge Graphs (KGs) can partially mitigate this issue, existing methods primarily treat KGs as static knowledge repositories, overlooking the critical disparity between KG and LLM knowledge, and failing to fully exploit the reasoning capabilities inherent in KGs. To address these limitations, we propose Pyramid-Driven Alignment (\pa), a novel framework for seamlessly integrating LLMs with KGs. \pda utilizes Pyramid Principle analysis to construct a hierarchical pyramid structure. This structure is designed to reflect the input question and generate more validated deductive knowledge, thereby enhancing the alignment of LLMs and KGs and ensuring more cohesive integration. Furthermore, \pda employs a recursive mechanism to harness the underlying reasoning abilities of KGs, resulting in more accurate knowledge retrieval for question-answering tasks. Our experimental results reveal a substantial performance advantage of \pda over state-of-the-art baselines, with improvements reaching 26.70\% and 26.78\%.
\end{abstract}

%

\section{Introduction}
Large language models (LLMs) exhibit remarkable intrinsic reasoning capabilities that enable them to excel in complex inferential tasks across a variety of natural language processing tasks \cite{feng_language_2023,kojima_large_2023,brown_language_2020}. However, LLMs often encounter challenges when presented with queries that necessitate specialized expertise exceeding their internal knowledge. Furthermore, their tendency to fabricate information, a phenomenon known as hallucination, significantly compromises their reliability and trustworthiness \cite{bang_multitask_2023,li_dawn_2024}. Previous research has demonstrated that integrating LLMs with the structured, clearly defined and explainable knowledge provided by knowledge graphs (KGs) can effectively address the dual challenges of hallucination and knowledge limitations. By grounding LLM reasoning in factual, verifiable data, KG offers a robust framework for enhancing the accuracy, reliability, and comprehensiveness of LLM-generated outputs. The synergistic integration of LLMs and KGs has emerged as a focal point in contemporary research and applications\cite{huang_survey_2023,agrawal_can_2024,gao_retrieval-augmented_2024,structgpt,keqing,baek2023knowledge}. 

\begin{figure}
    \centering
    \includegraphics[width=1\linewidth]{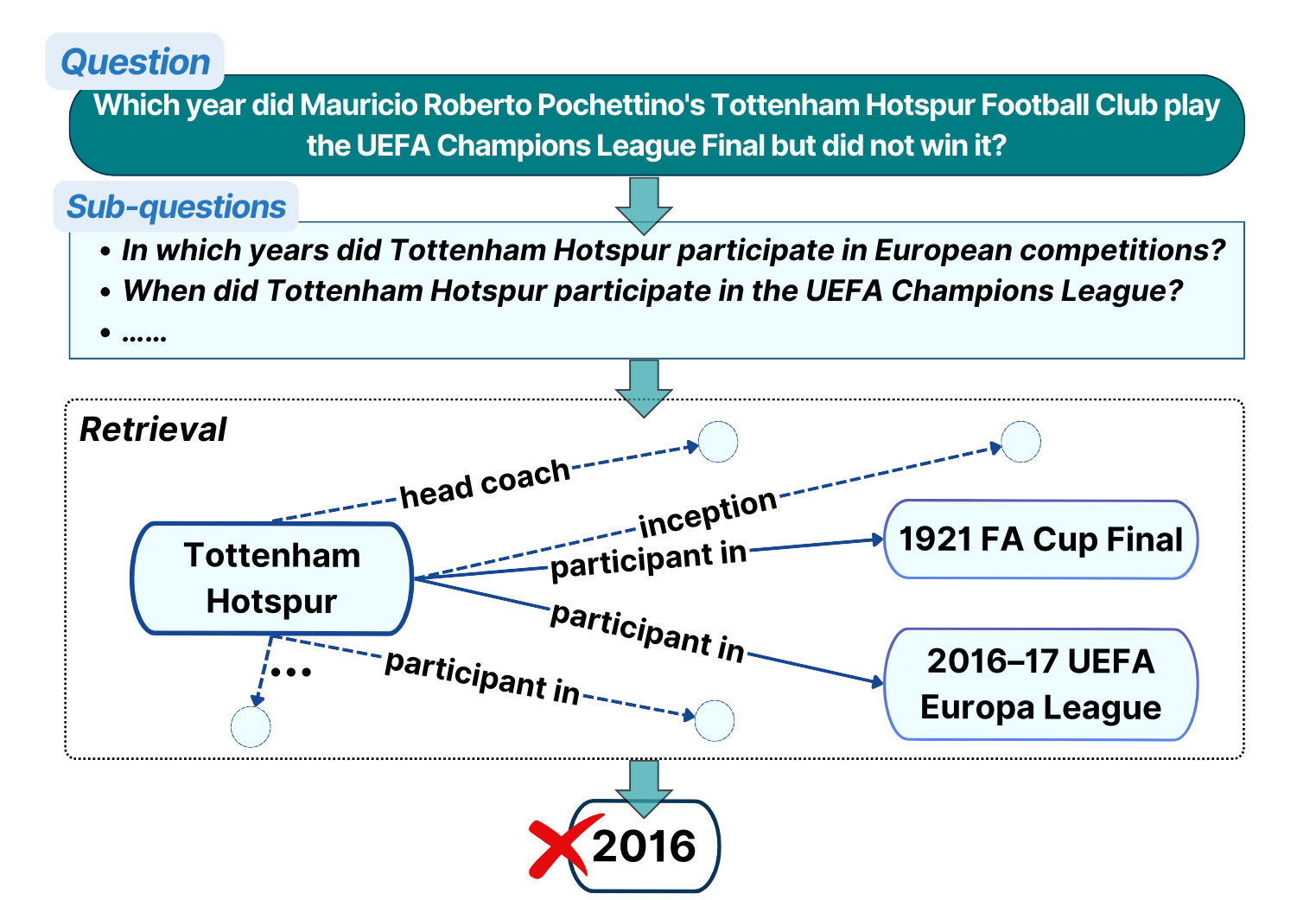}
    \caption{An issue of the knowledge generated in LLMs reasoning and how it leads to irrelevant knowledge retrieval and subsequent errors in the final answer.}
    \label{fig:intro-pic}
\end{figure}

Existing methods for integrating LLMs and KGs typically follow two primary paradigms. The first directly retrieves relevant triples in KG based on query entities \cite{wang2024boostinglanguagemodelsreasoning,baek2023knowledge}. The second approach focuses on guiding LLMs to generate a series of relevant knowledge, such as question decomposition or sub-answer generation \cite{keqing,park2024graphelicitationguidingmultistep}, to identify corresponding triples. However, both methods often fall short in bridging the knowledge gap between LLMs and KGs, and underutilize KG's inherent reasoning potential. As illustrated in Figure \ref{fig:intro-pic}, consider the question, \textit{Which year did Mauricio Roberto Pochettino's Tottenham Hotspur Football Club play the UEFA Champions League Final but did not win it}? This is a complex multi-hop question that requires reasoning across several pieces of information. In this scenario, LLM attempts to break the question down into multiple sub-questions, as shown in Figure \ref{fig:intro-pic}. It then retrieves relevant knowledge from KG based on these sub-questions, ultimately synthesizing this information to construct the final answer.

However, this approach, which solely relies on LLM-generated knowledge, neglects the deductive nature of KG reasoning. By disregarding logical interconnections between knowledge, this method increases the risk of retrieving irrelevant or misleading information from KG, potentially compromising response accuracy \cite{wu-etal-2024-decot}. For instance, sub-questions like \textit{When did Tottenham Hotspur participate in the UEFA Champions League?} can inadvertently retrieve irrelevant triples such as \textit{(Tottenham Hotspur, participated in, 1921 FA Cup Final)} or \textit{(Tottenham Hotspur, participated in, 2016–17 UEFA Europa League)}. These misleading triples can divert the LLM from the correct reasoning path, resulting in incorrect answers like \textit{2016}, instead of the correct \textit{2019}. As illustrated in Figure \ref{fig:intro-pic}, effectively addressing complex, multi-hop questions requires fully integrating LLMs and KGs capabilities \cite{oda}. Aligning LLMs outputs with the unique characteristics of KGs enhances this integration. However, achieving this alignment poses several challenges. First, how can we guide LLMs to generate knowledge that aligns with KG knowledge? Second, how can we leverage the aligned knowledge from LLMs to effectively utilize the inherent reasoning capabilities of KGs, thereby optimizing the integration of LLMs and KGs?



To address these challenges, we propose Pyramid-Driven Alignment (\pa), an innovative approach designed to align LLM and KG knowledge and optimize the integration of their reasoning capabilities. In \pa, we first design an effective reasoning method based on the Pyramid Principle \cite{minto2021} to align with KG knowledge. Specifically, \pda uses the 5W1H framework \cite{han20205w1hbasedexpressioneffectivesharing} to reflect the question, and organizes the results into a hierarchical pyramid structure. This approach enables \pda to derive validated deductive knowledge at the base of the pyramid (see Figure \ref{fig:workflow}), ensuring alignment with the deductive reasoning paths in KG. After establishing this Pyramid alignment, \pda employs a recursive mechanism to fully unlock reasoning capabilities of KG, ensuring efficient and accurate retrieval of task-relevant knowledge in KG. By integrating these elements, \pda effectively harmonizes the reasoning capabilities of both KG and LLM, optimizing their combined effectiveness.


To rigorously evaluate our proposed method, we conducted comprehensive experiments across three challenging datasets: 2WikiMultihopQA \cite{2WikiMultihopQA}, Mintaka \cite{sen-etal-2022-mintaka}, and WebQuestionsSP (WebQSP) \cite{WebQuestionsSP}. Our approach consistently surpassed state-of-the-art baselines, demonstrating its effectiveness in addressing complex question-answering tasks. Specifically, we achieved accuracy improvements of up to 19.00\% and 26.70\% on Mintaka, and 27.68\% and 26.78\% on WebQSP when utilizing GPT-3.5 and GPT-4o mini, respectively. The key contributions of this work are outlined below:
\begin{itemize}
    \item We propose a novel method named \pda to enhance the integration of LLMs and KGs. By pioneering the application of the Pyramid Principle to LLMs, \pda generates deductive knowledge that aligns with KG knowledge, facilitating seamless LLM-KG integration.

    
\end{itemize}
\begin{itemize}
    \item We design a recursive mechanism that leverages KG's inherent reasoning capabilities to improve LLM-KG integration and extract more accurate knowledge. By combining this with LLM's reasoning abilities, \pda significantly boosts question-answering performance.

    
\end{itemize}
\begin{itemize}
    \item We conducted experiments across three datasets and achieved state-of-the-art (SOTA) performance, showcasing the robustness and effectiveness of our approach.
\end{itemize}
\section{Methods}
\begin{figure*}[t]
    \centering
    \includegraphics[width=1\linewidth]{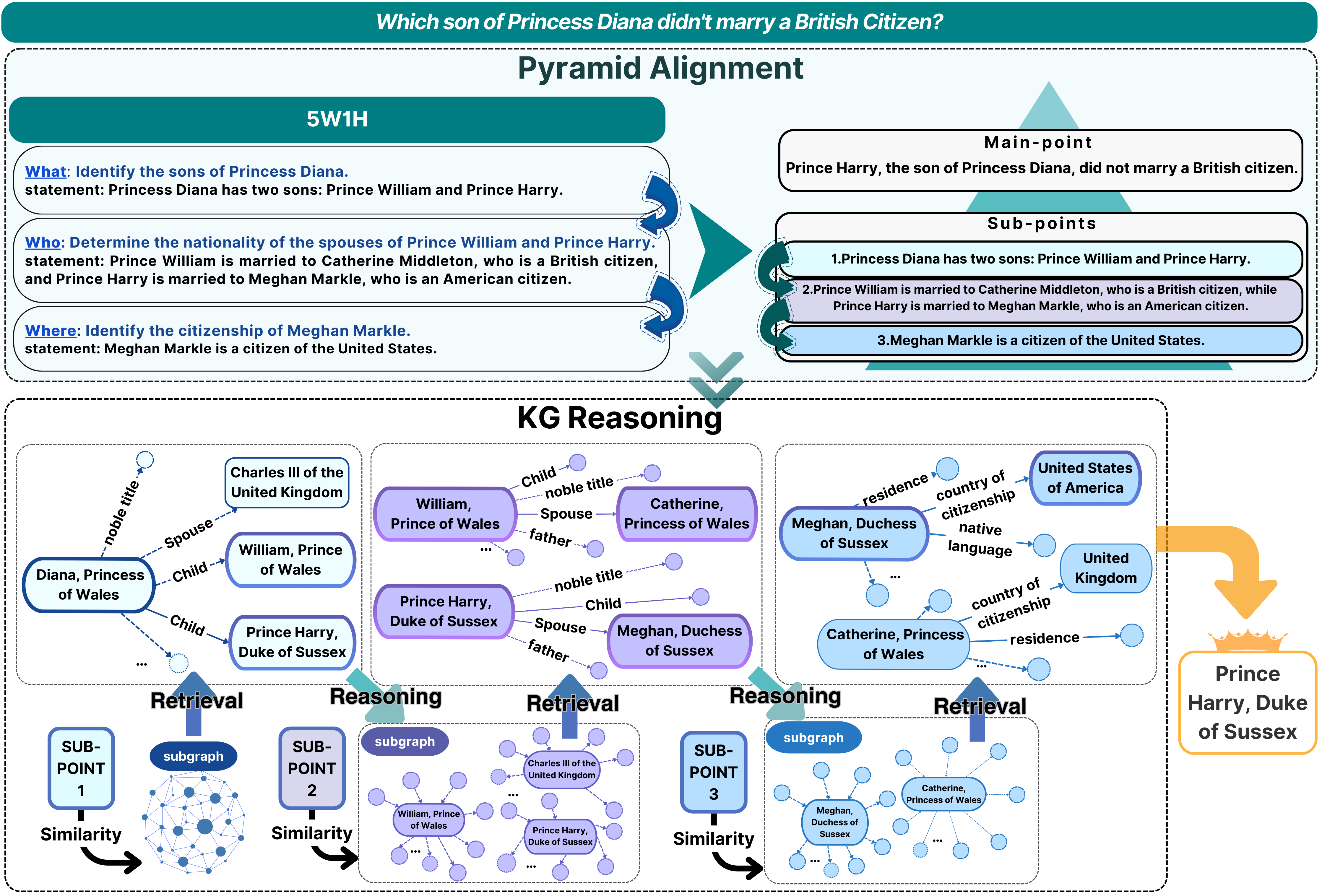}
    \caption{The overall framework of \pa. Pyramid Alignment: given a question, we apply the 5W1H framework to reflect it and construct a hierarchical pyramid that generates deductive knowledge. This deductive knowledge facilitates the alignment between LLM and KG. KG Reasoning: to utilizing the alignment knowledge, we leverage the inherent reasoning capabilities of KG to retrieve more accurate triples for answering the given question. Finally, we prompt LLM to generate the answer using these retrieved triples.}
    \label{fig:workflow}
\end{figure*}

For KG-driven tasks, effectively integrating the reasoning capabilities of LLM and KG is crucial for accurate knowledge acquisition, especially in question answering. Given a natural language question \( q \in Q \) and a Knowledge Graph \( G \), the task \( T \) is defined as generating an answer \( a \in A \), as follows:
\[
T : (G, q) \rightarrow a
\]
In this work, the proposed \pda prompts the LLM based on the Pyramid Principle to generate deductive knowledge, aligning the knowledge of LLM and KG. Building upon this alignment, \pda further harnesses the intrinsic reasoning capabilities of KG to extract more precise and relevant knowledge for the target task \( \tau \). \pda  comprises the following two components as illustrated in Figure \ref{fig:workflow}:
\begin{itemize}
    \item \textbf{Pyramid Alignment}: This module, denoted as \( P \), applies the Pyramid Principle to construct a set of deductive knowledge that aligns with the reasoning paths of KG. This alignment ensures a seamless integration between LLM and KG, empowering the model to excel at complex question-answering tasks.
    

    \item \textbf{KG Reasoning}: To leverage the inherent reasoning capabilities of KG, this module \( K \), iteratively refines knowledge acquisition. In each iteration \( i \), it retrieves a more accurate set of triples \(\mathcal{T}_i = \{(h_i, \rho_i, t_i) \mid h_i \in E, t_i \in E, \rho_i \in R\}\), (where \( E \) is the set of entities and \( R \) is the set of relations) by incorporating alignment knowledge \(I_i\). This synergy harnesses the strengths of both LLM and KG, enhancing effectiveness in tackling complex tasks.


\end{itemize}

\subsection{Pyramid Alignment}

By strategically employing the Pyramid Principle, the Pyramid Alignment module meticulously constructs deductive knowledge $\mathcal{I}$ aligning with KG's intricate reasoning paths. This process can be formally expressed as:
\[
    I = P(q)
\]
To leverage LLM capabilities, the given question \( q \) is reflected into its constituent 5W1H elements, with each element denoted as \( \mathcal{W}_i \mid i \in \mathcal{Z} \), where \( \mathcal{Z} \) represents the set of the relevant 5W1H elements determined by the LLM, with a cardinality of \(n\). Each \( \mathcal{W}_i \) is deductively related to the others.
\[
    \mathcal{W} = \text{5W1H}(q), \quad \mathcal{W} = \{\mathcal{W}_i \mid i \in \mathcal{Z}\}
\]
Subsequently, \pda generates statements for each $\mathcal{W}_i$, serving as foundational evidence for constructing a hierarchical pyramid structure.
\[
    S = \text{Statement}(\mathcal{W}, q) \quad \mathcal{S} = \{\mathcal{S}_i \mid i \in \mathcal{Z}\}
\]
Leveraging the statements $ S $ and the pyramid hierarchical structure, \pda generates a main point $\mathbf{M}$ and corresponding sub-points $\mathcal{I}$.
\[
    \mathcal{M} = \text{MainPoint}(S)
\]
\[
    I = \text{SubPoints}(\mathbf{M}, S) \quad \mathcal{I} = \{\mathcal{I}_i \mid i \in \mathcal{Z}\}
\]
These sub-points are the knowledge from LLM which interconnect through deductive reasoning, dynamically adjusted based on the influence of each $S_i$ and $\mathcal{M}$. This adaptive approach ensures that the last sub-point is directly relevant to answering the question.

When applying the 5W1H framework to reflect a given question, LLM may face ambiguities with certain entities in the generated sub-points, potentially disrupting deductive reasoning. To address this, the Pyramid Alignment module prompts LLM to use entity types from the Stanford Named Entity Recognizer \cite{finkel-etal-2005-incorporating} to disambiguate these entities within each sub-point. By substituting unknown entities with their corresponding entity types, we fortify the Pyramid Alignment module's robustness, concurrently diminishing the negative influence of these entities on the downstream KG Reasoning module.

These deductive sub-points as alignment knowledge are sent to the next module. All strategies in the Pyramid Alignment module are implemented within a single input prompt, detailed in the Appendix.

\subsection{KG Reasoning}

This module leverages both the alignment knowledge from the Pyramid Alignment module and the inherent reasoning capabilities of KG for retrieving more accurate knowledge. At each iteration \(i\), \pda utilizes the entities \(E_i\), the deductive knowledge \(\mathcal{I}_i\) generated by the Pyramid Alignment module, and the subgraph \(G_{i-1}\) reasoned from the previous iteration to produce a set of triples \(\mathcal{T}_i\). This process is as below:
\[
    \mathcal{T}_i = K([I_i,E_i,G_{i-1}])
\]
The output of the KG Reasoning module, denoted as \( \mathcal{T}\), is obtained by summing the set of triples generated in each iteration:
\[
\mathcal{T} = \sum_{i=1}^{j} \mathcal{T}_i
\]

To achieve a more effective integration of LLM and KG, it is crucial to harness the inherent reasoning capabilities of KG. To this end, we propose a recursive strategy. This strategy consists of two main components: retrieval and reasoning. In the retrieval step, alignment knowledge from the Pyramid Alignment module is used to obtain more accurate triples from KG. In the reasoning step, these triples are used to build a more focused subgraph for the next iteration. Leveraging task-relevant entities \(E_q = \{e_{q0},\ldots,e_{qm}\}, \text{ where } m \text{ is the number of elements}\) derived from those embedded within the question \(q\), the KG Reasoning module initializes the subgraph as \(G_0\). The process then iterates through retrieval and reasoning steps for each iteration \(i\) until the final index in $\mathcal{Z}$ is reached. The details are follows:
\begin{algorithm}[htbp]
\caption{KG Reasoning Algorithm}
\textbf{Require:} Question \(q\), alignment knowledge $ \mathcal{I} $ and limit \(n\).
\begin{algorithmic}
\STATE Initialize task-relevant entities \(E_q = \{e_{q0}, e_{q1}, \ldots, e_{qm}\}\) with the entities in \(q\)
\STATE Initialize subgraph \(G_0\)
\FOR{iteration \(i\) from 1 to \(n\)}
    \FOR{\( (h, r, t) \in  G_{i-1} \)}
        \STATE Cosine Similarity (\(\mathcal{I}_i, h, r, t)\)
    \ENDFOR
    \STATE Sort all the triples according to the similarity scores
    \STATE Select top \(N\) triples and denote as \(\mathcal{T}_i\)
    
    \STATE Construct \(G_i\) with the neighboring triples of  \(t \mid t \in\mathcal{T}_i\)
    \STATE Set \(G_i\) as input for the next iteration
\ENDFOR
\STATE Output the set of triples \(\mathcal{T} = \sum_{i=1}^{n} \mathcal{T}_i\)
\end{algorithmic}
\end{algorithm}

\noindent\textbf{Retrieval}

\begin{itemize}
    \item Convert each triple in $G_{i-1}$ into text format as $h, r, t$.
    \item Compute the embedding similarity between the current alignment knowledge $\mathcal{I}_i$ and all $h, r, t$ in $G_{i-1}$ using the following formula:
    \[
    \text{Cosine Similarity}(\mathbf{v}_{\mathcal{I}_i}, \mathbf{v}_{h,r,t}) = \frac{\mathbf{v}_{\mathcal{I}_i} \cdot \mathbf{v}_{h,r,t}}{\|\mathbf{v}_{\mathcal{I}_i}\| \|\mathbf{v}_{h,r,t}\|}
    \]
    \item Sort all triples in descending order of similarity.
    \item Select the top N triples $\mathcal{T}_i$ to update the output of KG reasoning process, denoted as $\mathcal{T}$.
\end{itemize}

\noindent\textbf{Reasoning}

\begin{itemize}
    \item The triples retrieved in the previous step, \( \mathcal{T}_i \), are used for reasoning.
    \item The tail entities \( t \mid t \in \mathcal{T}_i \) are extracted to construct a more focused subgraph for the current iteration.
    \item Neighboring triples of each tail entity are retrieved from KG to form the focused subgraph, denoted as \( G_i \).
    \item This subgraph \( G_i \) is then used in the retrieval process for the next iteration.
\end{itemize}

Upon comprehensive collection of all triples $\mathcal{T}$, these are integrated with the given question and presented to the LLM for response generation. This enriched input facilitates the production of a significantly more accurate and informative answer. 


\section{Experiments}
\subsection{Dataset}
To assess the efficacy of our \pda approach for complex question-answering tasks, we employ three KBQA datasets which necessitate advanced reasoning capabilities: 2WikiMultihopQA, Mintaka, and WebQSP. The details of the three KBQA datasets are illustrated in Table \ref{tab:dataset}.

\begin{table}[h]
\centering
\begin{tabular}{lcccc}
\toprule
Dataset & Train & Test & Type \\
\midrule
WebQSP         & 27,734           & 1,639             & Multi-hop     \\ 
Mintaka             & 14,000          & 4,000            & Various    \\ 
2WikiMultiHopQA      & 154,878          & 12,576            & Multi-hop          \\ 
\bottomrule
\end{tabular}
\caption{The dataset statistics}
\label{tab:dataset}
\end{table}

WebQSP is built upon the Freebase KG \cite{bollacker2008freebase}. This dataset consists of 1,639 test instances, which we use to evaluate the performance of \pa. Mintaka, a more recent dataset designed for complex KGQA tasks, is based on the Wikidata knowledge graph. While Mintaka includes questions in eight languages, our study focuses exclusively on the English test set. Due to computational constraints, we randomly selected 1,000 instances from the Mintaka English test set for evaluation. 2WikiMultiHopQA, designed for multi-hop question answering, was also included in our experiments. To ensure comparable dataset sizes, we randomly sampled 1,000 examples from this dataset. We employed Hits@1 \cite{blagec2020critical} accuracy as the primary evaluation metric.

\subsection{Setup}
We employed the GPT-3.5 Turbo (`GPT-3.5' is used to refer to this throughout the paper) and GPT-4o mini model\footnote{\url{https://platform.openai.com/docs/models}} for our experiments. To ensure consistency and comparability, the model's temperature parameter was fixed at 0.4, and the maximum token length was constrained to 1,000 tokens throughout all experimental runs. For the KG reasoning module, we configured key parameters based on the setup in KAPING \cite{baek2023knowledge}, setting the top $N$ triples to 10.

To facilitate information retrieval from the Wikidata KG, we leveraged the capabilities of the \textit{simple-wikidata-db} Python library\footnote{\url{https://github.com/neelguha/simple-wikidata-db}}. This library provides a comprehensive toolkit for managing large-scale knowledge graphs, encompassing functionalities such as downloading the latest Wikidata dump, efficient staging, and distributed query execution. The Wikidata dump was distributed across a cluster of five AWS EC2 instances, each with 48 cores and 768GB of memory. The smaller Freebase dump was hosted on a single machine with 48 cores and 256GB of memory, as specified in the Think-on-Graph (ToG) setup \cite{sun_ToG}. 

The Freebase dump is managed using OpenLink Virtuoso, a versatile database supporting relational, RDF, and graph data models. Although Virtuoso is widely used for KG applications, its default configuration led to prohibitively long loading times for Freebase, nearly two days with the ToG setup. After careful parameter optimization, we reduced the loading time to under 30 minutes. Detailed configuration parameters are provided in the Appendix.

\subsection{Baseline Models}
To rigorously evaluate the effectiveness and capabilities of PDA in handling complex question-answering tasks, we compared it against several SOTA methods designed to address the same tasks. Our comparative analysis categorizes the SOTA methods into two major groups: prompt-based methods and knowledge-augmented methods. 
Prompt-based methods, such as CoT \cite{kojima_large_2023} and Self-Consistency \cite{self-consistency}, rely exclusively on the inherent reasoning abilities of LLMs without incorporating external knowledge. For a fair comparison, we replicated CoT and Self-Consistency using GPT-3.5 and GPT-4o mini, strictly following the original implementation guidelines. On the other hand, knowledge-augmented methods, such as KAPING \cite{baek2023knowledge}, integrate KG with LLM reasoning to effectively tackle question-answering tasks. We also reproduced KAPING with both GPT-3.5 and GPT-4o mini models to ensure a fair comparison. Additionally, two KG-driven methods, Subgraph-KGQA \cite{salnikov2023large} and Knowledge-Driven CoT \cite{Knowledge-driven-cot}, were included in the comparison.

\subsection{Main Results}
\begin{table*}[ht]
\centering
\small
\begin{tabular}{@{}lcccc@{}}
\toprule
Method & 2WikiMultiHopQA & Mintaka & WebQSP & Average \\
\midrule
\multicolumn{5}{c}{\textit{w.o. Knowledge}} \\
\midrule
CoT w/GPT-3.5 & 28.20 & 64.30 & 38.56 & 43.69 \\
CoT w/GPT-4o mini & 28.10 & 63.90 & 38.62 & 43.54 \\
\midrule
Self Consistency w/GPT-3.5 & 10.90 & 60.50 & 54.97 & 42.12 \\
Self Consistency w/GPT-4o mini & 15.20 & 61.90 & 56.01 & 44.37 \\
\midrule
\multicolumn{5}{c}{\textit{w.t. Knowledge}} \\
\midrule
KAPING w/GPT-3.5 & 26.60 & 46.70 & 45.57 & 39.62 \\
KAPING w/GPT-4o mini & 30.40 & 46.50 & 52.03 & 42.98 \\
\midrule
KG-driven SOTA w/GPT-3.5 & - & 33.00\textsuperscript{1} & 68.60\textsuperscript{2} & 50.80 \\
\midrule
PDA (Ours) w/GPT-3.5 & \textbf{33.70} & \textbf{65.70} & \textbf{73.25} & \textbf{57.55} \\
PDA (Ours) w/GPT-4o mini & \textbf{35.90} & \textbf{73.20} & \textbf{78.81} & \textbf{61.90} \\
\bottomrule
\end{tabular}
\caption{Performance Comparison of different methods on 2WikiMultiHopQA, Mintaka, and WebQSP. Bold scores represent the best performance among all methods using each corresponding model. The KG-driven SOTA includes: 1: Subgraph-KGQA \cite{salnikov2023large}, 2: Knowledge-Driven CoT \cite{Knowledge-driven-cot}.}
\label{tab:main}
\end{table*}

As indicated in Table \ref{tab:main}, our \pda consistently outperforms existing approaches on the 2WikiMultiHopQA, Mintaka, WebQSP datasets. Compared to best-performing prompt-based methods operating without external knowledge, \pda achieves significant improvements, with gains of 5.50\% (2WikiMultiHopQA), 1.40\% (Mintaka), and 17.24\% (WebQSP) using GPT-3.5, and 7.80\% (2WikiMultiHopQA), 9.30\% (Mintaka), and 22.80\% (WebQSP) using GPT-4o mini. Furthermore, when compared to KG-driven SOTA, \pda also demonstrates superior performance. \pda surpasses Subgraph-KGQA by 32.70\% on the Mintaka dataset, and Knowledge-Driven CoT by 4.65\% on WebQSP. Against KAPING, \pda exhibits substantial gains of 7.10\%, 19.00\%, and 27.68\% on 2WikiMultiHopQA, Mintaka, and WebQSP, respectively, when using GPT-3.5, while 5.50\%, 26.70\%, and 26.78\%, respectively, with GPT-4o mini. These results underscore the effectiveness of \pda in leveraging the combined reasoning capabilities of LLM and KG, leading to substantial improvements for complex question-answering tasks.

To demonstrate the efficacy of \pa, we examine a specific case as illustrated in Figure \ref{fig:workflow}. In the Pyramid Alignment module, the question \textit{Which son of Princess Diana didn't marry a British Citizen?} is methodically reflected through the 5W1H framework. This process initiates by identifying \textit{What—Identify the sons of Princess Diana} supported by the statement \textit{Princess Diana has two sons: Prince William and Prince Harry}. It then deductively progresses to \textit{Who—Determine the nationality of the spouses of Prince William and Prince Harry}, with the statement \textit{Prince William is married to Catherine Middleton, a British citizen, while Prince Harry is married to Meghan Markle, an American citizen}. Finally, it addresses \textit{Where-Identify the citizenship of Meghan Markle} with the statement \textit{Meghan Markle is a citizen of the United States}. Leveraging this analysis, the Pyramid Principle is employed to establish the main point and the deductive sub-points including \textit{Princess Diana has two sons: Prince William and Prince Harry}, \textit{Prince William is married to Catherine Middleton, a British citizen, while Prince Harry is married to Meghan Markle, an American citizen}, and \textit{Meghan Markle is a citizen of the United States}. These three sub-points as alignment knowledge are sent to KG Reasoning module, where the inherent reasoning capabilities of KG are effectively utilized. 

For the initial iteration of the KG Reasoning module, \pda initiates the retrieval process by using the first sub-point, \textit{Princess Diana has two sons: Prince William and Prince Harry}. Leveraging this, the module retrieves relevant triples from the KG, such as \textit{(Diana, Princess of Wales, child, William, Prince of Wales)} and \textit{(Diana, Princess of Wales, child, Prince Harry, Duke of Sussex)}. These triples is collected and used to reason a more focused subgraph for the next iteration. In the second iteration, the \pda applies the second sub-point, \textit{Prince William is married to Catherine Middleton, who is a British citizen, while Prince Harry is married to Meghan Markle, who is an American citizen} to retrieve relevant triples, such as \textit{(William, Prince of Wales, spouse, Catherine, Princess of Wales)} and \textit{(Prince Harry, Duke of Sussex, spouse, Meghan, Duchess of Sussex)}. The focus here is on confirming the identity of their spouses, ensuring that \pda correctly associates each prince with their respective partner, refining the collection for more accurate knowledge necessary to answer the question. Additionally, the tail entities from the retrieved triples, such as \textit{Meghan, Duchess of Sussex} and \textit{Catherine, Princess of Wales}, are then employed to reason a more focused subgraph for the next iteration. In the final iteration, the sub-point \textit{Meghan Markle is a citizen of the United States} retrieves the final triples, such as \textit{(Meghan, Duchess of Sussex, country of citizenship, United States of America)}. The aggregated triples from all iterations are then provided to the LLM, ultimately confirming the correct answer: \textit{Prince Harry, Duke of Sussex}. 

To further illustrate the versatility and effectiveness of \pa, we present a case study addressing the question: \textit{What is the name of the capital city of the country that only shares a border with Spain}? In the Pyramid Alignment module, the 5W1H framework begins with identifying \textit{What—Identify the country that shares a border only with Spain}. The statement supporting this is \textit{The country that shares a border only with Spain is Andorra}. This leads to the next deductive element, \textit{Where—Determine the capital city of Andorra}, substantiated by the statement generated from the LLM \textit{The capital city of Andorra is Andorra la Vella}. Using this deductive process, the Pyramid Principle establishes the main point: \textit{The capital city of the country that only shares a border with Spain is Andorra la Vella}. This main point is reinforced by the deductive sub-points derived from the LLM, including \textit{The country that shares a border only with Spain is Andorra} and \textit{The capital city of Andorra is Andorra la Vella}, corresponding to the \textit{What, Where} aspects of the 5W1H framework. 

Utilizing the sub-points as alignment knowledge in the KG Reasoning module, \pda initiates the retrieval process by using the first sub-point \textit{The country that shares a border only with Spain is Andorra}. Although this sub-point is incorrect due to hallucinations in the LLM, \pda still identifies and collects validated triples such as \textit{(Spain, shares border with, Portugal)}, \textit{(Spain, shares border with, Andorra)}, and \textit{(Andorra, shares border with, France)}. Leveraging these validated triples, \pda can still reason a validated subgraph for the next iteration. In the subsequent iteration, \pda uses the second sub-point, \textit{The capital city of Andorra is Andorra la Vella}, to retrieve more accurate triples like \textit{(Portugal, capital, Lisbon)} and \textit{(Andorra, capital, Andorra la Vella)}. This retrieval correctly identifies \textit{Lisbon} as the capital of the country that only shares a border with Spain, effectively correcting the earlier hallucination. This case vividly demonstrates the resilience of \pa. Even though the deductive knowledge generated in the Pyramid Alignment module contained the erroneous knowledge, \pda harnesses reasoning capabilities of KG to extract accurate factual triples, mitigating the impact of the hallucination. 

These case studies clearly illustrate how \pda leverages alignment knowledge reasoned from LLM, and the inherent reasoning capabilities of KG to improve the retrieval of factual triples, therefore delivers robust and accurate answers and effectively address complex question-answering tasks. These case studies can be found in the Appendix.


\section{Discussion}
To assess the individual contributions of Pyramid Alignment and KG Reasoning modules to our \pda approach, we conducted a series of ablation experiments on the WebQSP test dataset using the GPT-4o mini model. For this analysis, we randomly collected 400 samples from the WebQSP test dataset. A comparison of performance across different configurations is presented in Table \ref{tab:ablation-comparison}.

We conducted experiments with the model excluding both Pyramid Alignment and KG Reasoning modules, as well as separate experiments with each module removed individually. Ablation studies reveal the critical roles of both Pyramid Alignment and KG Reasoning in \pa. Excluding the Pyramid Alignment diminishes performance to 78.00\% from 81.75\%, emphasizing its role in structuring information and enhancing overall reasoning. Removing KG Reasoning leads to a performance of 79.50\%, indicating KG's contribution to reasoning through structured external knowledge. Removing both components drastically reduces performance to 72.00\%, highlighting their synergistic impact. \pa’s integration of Pyramid Alignment and KG Reasoning achieves optimal performance at 81.75\%, demonstrating the effectiveness of combining LLM and KG reasoning capabilities.

\begin{table}[h]
\centering
\begin{tabular}{lc}
\toprule
Method & WebQSP \\
\midrule
Without Pyramid Alignment & 78.00 \\
Without KG reasoning & 79.50 \\
Without both & 72.00 \\
PDA & \textbf{81.75} \\
\bottomrule
\end{tabular}
\caption{Ablation Comparison on WebQSP}
\label{tab:ablation-comparison}
\end{table}

\subsection{Effect of Pyramid Alignment and KG Reasoning}

To investigate the impact of excluding both Pyramid Alignment and KG Reasoning, we examine two specific cases where the model excludes both Pyramid Alignment and KG Reasoning modules.

In the first scenario, the question posed is \textit{Where was Franz Ferdinand from}? Without incorporating Pyramid Alignment and KG Reasoning, the LLM generated several pieces of evidences: \textit{Franz Ferdinand was a member of the Habsburg dynasty}, \textit{He was born in the city of Graz, which is located in present-day Austria}, and \textit{Franz Ferdinand held the title of Archduke of Austria}. Based on these information, the model retrieved incorrect triples and concluded the answer \textit{Austria}. This case illustrates that the absence of Pyramid Alignment and KG Reasoning leads to invalid knowledge from LLM and the retrieval of irrelevant information from KG, ultimately resulting in inaccurate answers.

In the second scenario, the question is \textit{When did Annie open}? The LLM produced knowledge including \textit{Annie refers to a specific establishment or event, such as a restaurant or a store}, \textit{Annie is located in a specific city or area}, \textit{The owner or founder of Annie is a specific individual or entity}, and \textit{Annie was opened through a specific process or event}. 
The retrieved top triples based on the LLM knowledge were: \textit{(Musical theatre, theater.genre.plays, From the Second City)} and \textit{(unnamed\_entity, theater.production.venue, Annie)}. In this example, the model retrieved many triples containing unnamed entities. This issue arises because when the reasoning of LLM is not robust, it may generate many uncertain pieces of evidence, such as \textit{The owner or founder of Annie is a specific individual or entity}. When retrieving information from KG, particularly from an outdated knowledge graph like Freebase, which contains a certain amount of unnamed entities \cite{sun_ToG}, there is a high likelihood of retrieving these unnamed entities, leading to erroneous answers. In contrast, our \pda provides a more reliable reasoning approach by addressing the influence of unknown entities through their types. Furthermore, it leverages the inherent reasoning capabilities of KG to ensure the retrieval of validated triples, resulting in a more robust integration with various KGs.



\subsection{Performance across Different Backbone Models}
\begin{table}[h]
\centering
\begin{tabular}{lc}
\toprule
Model & WebQSP \\
\midrule
GPT-4o mini & 81.75 \\
DeepSeek-V2 & 78.75 \\
\bottomrule
\end{tabular}
\caption{Performance comparison using different backbone models on WebQSP}
\label{tab:backbone-comparison}
\end{table}

To evaluate the generalizability of our \pda across different LLM architectures, we assessed its performance on the WebQSP dataset using GPT-4o mini and DeepSeek-V2 which is a cost-effective mixture-of-experts language model, outperforms the LLaMA3 70B Instruct model on standard benchmarks \cite{bi2024deepseek}. As presented in Table \ref{tab:backbone-comparison}, \pda consistently delivers strong results across both models. Specifically, \pda paired with GPT-4o mini demonstrates excellent performance, achieving an accuracy of 81.75\%, while using DeepSeek-V2 results in an accuracy of 78.75\%.

\section{Related Works}

\subsection{Reasoning with LLM Prompting}
Recent research has focused on enhancing LLM reasoning through prompting. The Chain-of-Thought (CoT) method \cite{CoT,kojima_large_2023} has been particularly influential, leading to advancements such as Complex-CoT \cite{fu_complexity-based_2023}, Tab-CoT \cite{jin2023tab}, Self-Consistency \cite{self-consistency}, Tree of Thoughts (ToT) \cite{yao_tree_2023}, and Graph of Thoughts (GoT) \cite{besta_graph_2024}. Additionally, studies have explored deductive reasoning in LLMs, including \citet{ling2024deductive,zhu_deductive_2024,liu_concise_2024,holmes2023evaluating,yang2023logical}, and \citet{huang2022towards}. Furthermore, incorporating external documents into prompting for reasoning has been investigated, as demonstrated in \citet{li2023leveraging} and \citet{shi2024generate}. However, they solely focus on the reasoning capabilities of LLMs, without adequately addressing the issue of hallucinations that LLMs may produce.

\subsection{KG-Augmented LLM}
KG-augmented LLM inference leverages KG for real-time knowledge updates during inference without retraining \cite{pan2024unifying}. Existing methods primarily focus on transforming structured KG into textual representations for LLM. These approaches vary in their methodologies, including converting triples into sentence-like formats \cite{li2023graph,luo2023chatrule} and linearizing question-specific subgraphs such as \citet{salnikov2023large}, Retrieve-Rewrite-Answer \cite{wu2023retrieve} and Chain-of-knowledge \cite{li2023chain} as refinement for better task-solving~\cite{tao2024survey}.
Furthermore, several studies have integrated LLM reasoning with KG. \citet{li2024enhanced} present the approach by utilizing LLMs to selectively retrieve task-related subgraphs from KG. \citet{Knowledge-driven-cot} propose Knowledge-driven CoT, a question answering system incorporating CoT reasoning, while KAPING \cite{baek2023knowledge} and Keqing \cite{keqing} focus on prompt LLMs to reason and generate knowledge relevant to the question, these methods overlook the gap between LLM knowledge and KG knowledge. Recent advancements in retrieval-augmented methods have leveraged the structure of KG to enhance KGQA performance. RoG \cite{reasoning_on_graphs} enables LLMs to generate the knowledge aligned with paths structure of KG, while UniKGQA \cite{unikgqa} produces semantic matching features between questions and subgraph triples. 

Although these works integrate LLM and KG, they overlook the misalignment between LLM and KG knowledge, particularly regarding KG's reasoning paths. Additionally, they also fail to fully leverage the inherent reasoning capabilities of KG. As a result, their integration of LLM and KG does not reach its full potential. In contrast, our \pda aligns deductive knowledge with KG and synergistically combines the reasoning strengths of both LLM and KG. This approach leads to a more effective integration, enhancing the ability to address complex question-answering tasks.

\section{Conclusion}
In this work, we introduce a novel approach \pda for solve complex KBQA tasks. \pda innovatively employs a reasoning method based on the Pyramid Principle to generate deductive knowledge from LLM and aligns it with knowledge in KG. We also propose a recursive mechanism to fully harness the reasoning capabilities of KG, ensuring efficient and accurate retrieval of task-relevant knowledge. By synergistically combining the reasoning strengths of both LLM and KG, \pda effectively tackles complex question-answering tasks. Extensive experiments validate the effectiveness of \pa, showcasing its capability to handle intricate question-answering scenarios.

\bibliography{main}
\clearpage
\section{Technical Appendix}
\subsection{Case Study}\label{sec:case}
To demonstrate the effectiveness of \pda, we present several representative cases from the Mintaka dataset. We selected two examples to illustrate how \pda utilizes Pyramid Alignment and KG Reasoning modules to correctly address complex questions in Table \ref{tab:case1} and Table \ref{tab:case2}.

\begin{table}[htbp]
\centering
\begin{minipage}{\textwidth}
\begin{tabular}{|p{4cm}|p{11cm}|}
\hline

\textbf{Question} & \textit{What is the name of the capital city of the country that only shares a border with Spain?} \\
\hline

\textbf{Task-relevant Entities} & \textit{Q29: Spain} \\

\hline
\textbf{Pyramid Alignment} & \textbf{5W1H}: 
    \newline
    [
    \newline
    \textit{\textbf{What}}: Identify the country that shares a border only with Spain.
    \newline
    statement: The country that shares a border only with Spain is Andorra.
    \newline
    \newline
    \textit{\textbf{Where}}: Determine the capital city of Andorra.
    \newline
    statement: The capital city of Andorra is Andorra la Vella.
    \newline
    ]
    \newline
    \newline
    \textbf{Pyramid Principle}: 
    \begin{itemize}
    \item\textbf{Main-point}: 
    \newline
    The capital city of the country that only shares a border with Spain is Andorra la Vella.
    \end{itemize}
    
    \begin{itemize}
    \item\textbf{Sub-points}:
    \newline
    The country that shares a border only with Spain is Andorra.
    \newline
    The capital city of Andorra is Andorra la Vella.
    \end{itemize}

    \textbf{Thinking}: 
    \newline
    By identifying the country that shares a border only with Spain and confirming its capital, we establish that the capital city is Andorra la Vella.
    \newline
    \\
\hline
\textbf{KG Reasoning} & \textbf{Triples}: 
\begin{itemize}
\item\textbf{first iteration}: 
\newline
\textit{(Spain, shares border with, Portugal), (Portugal, shares border with, Spain), (France, shares border with, Benelux), (Spain, shares border with, Andorra), (Andorra, shares border with, France), (Andorra, shares border with, Spain)} 
\end{itemize}

\begin{itemize}
\item\textbf{second iteration}: 
\newline
\textit{(Andorra, capital, Andorra la Vella), (France, capital, Paris), (Portugal, capital, Lisbon)} 
\end{itemize}
\\
\hline
\textbf{Answer} & \textit{Lisbon} \\
\hline
\end{tabular}
\captionsetup{justification=centering}
\caption{Case 1}
\label{tab:case1}
\end{minipage}
\end{table}

\clearpage

\begin{table}[htbp]
\centering
\begin{minipage}{\textwidth}
\begin{tabular}{|p{4cm}|p{11cm}|}
\hline
\textbf{Question} & \textit{Who sang It Ain't Me Babe first, Johnny Cash or Bob Dylan?} \\
\hline

\textbf{Task-relevant Entities} & \textit{Q2706172: It Ain't Me Babe, Q42775: Johnny Cash, Q251309: Bob Dylan} \\

\hline
\textbf{Pyramid Alignment} & \textbf{5W1H}:
    \newline
    [
    \newline
    \textit{\textbf{What}}: Identify the original artist of the song "It Ain't Me Babe".
    \newline
    statement: Bob Dylan is the original artist of the song "It Ain't Me Babe".
    \newline
    \newline
    \textit{\textbf{When}}: Determine the release date of *It Ain't Me Babe* by Bob Dylan.
    \newline
    statement: Bob Dylan released "It Ain't Me Babe" in 1964.
    \newline
    \newline
    \textit{\textbf{Who}}: Identify who covered "It Ain't Me Babe" after Bob Dylan.
    \newline
    statement: Johnny Cash covered "It Ain't Me Babe" in 1965.
    \newline
    ]
    \newline
    \newline
    \textbf{Pyramid Principle}: 
    \begin{itemize}
    \item\textbf{Main-point}: 
    \newline
    Bob Dylan sang "It Ain't Me Babe" first, before Johnny Cash covered it.
    \end{itemize}
    
    \begin{itemize}
    \item\textbf{Sub-points}:
    \newline
    Bob Dylan is the original artist of the song "It Ain't Me Babe".
    \newline
    Bob Dylan released "It Ain't Me Babe" in 1964, while Johnny Cash covered it in 1965.
    \end{itemize}
    
    \textbf{Thinking}:
    \newline
    By identifying the original artist and the timeline of releases, we confirm that Bob Dylan was the first to sing *It Ain't Me Babe* before Johnny Cash.
    \newline
    \\
\hline
\textbf{KG Reasoning} & \textbf{Triples}: 
\begin{itemize}
\item\textbf{first iteration}: 
\newline
\textit{(``It Ain't Me Babe", composer, Bob Dylan), (``It Ain't Me Babe", lyrics, Bob Dylan), (``It Ain't Me Babe", performer, Bob Dylan)} 
\end{itemize}

\begin{itemize}
\item\textbf{second iteration}: 
\newline
\textit{(``It Ain't Me Babe", published in, Another Side of Bob Dylan), (``It Ain't Me Babe", recording date, +1964-06-09T00:00:00Z)}
\end{itemize}
\\
\hline
\textbf{Answer} & \textit{Bob Dylan} \\
\hline
\end{tabular}
\captionsetup{justification=centering}
\caption{Case 2}
\label{tab:case2}
\end{minipage}
\end{table}

\clearpage
\subsection{Prompt}\label{sec:prompt}
This section presents the prompt for Pyramid Alignment module in Table \ref{tab:PAprompt}.
\begin{table*}[htbp]
\centering
\begin{minipage}{\textwidth}
\begin{tabular}{{p{17cm}}}
\hline
\textbf{Prompt} \\ \hline
Using the Pyramid Principle to gather the facts that can answer the question: \%s. \\
Separate your response into the following parts: \\
5W1H: the analysis and reflection of the question based on 5W1H. \\
Main point: the main point of your response. \\
Sub-points: the facts that summarize the 5W1H analysis and support the main point. \\
\\
\textbf{Guidelines to Follow:} \\
First, use 5W1H to analyze and reflect on the question, and derive a main point. \\
- The 5W1H analysis should be step-by-step and deductive in a logical order. \\
- Only include the 5W1H elements that are relevant to the question. \\
- The statements should be factual based on each relevant 5W1H element. \\
- If you are unsure about a entity, you can use the entity type from the Stanford Named Entity Recognizer to replace the entity in the statement. \\
- Summarize the statements to derive the main point. \\
\\
Then, based on the 5W1H statements, provide sub-points that support the main point. \\
- Each sub-point should dialectically summarize a 5W1H statement. \\
- Each sub-point should follow a deductive relationship, meaning each sub-point should build upon or logically follow from the previous one. \\
- If you are unsure about a entity, you can use the entity type from the Stanford Named Entity Recognizer to replace the entity in the sub-points. \\
- Each sub-point should be a entity, not a suggestion, instruction, or procedure. \\
\\
\textbf{Rules:} \\
- If each sub-point includes multiple independent entity or information, separate them with a [SEP] token. \\
- Avoid giving instructions or explanations in the 5W1H statements and sub-points. \\
- If you do not know the entity, do not give suggestions, instructions, or explanations for why you do not know the entities. \\
- Think through the process step by step and list them out. \\
- Format your response in JSON as follows: \{"5W1H": [], "main-point": "", "sub-points": [], "thinking": ""\} \\
\\
\textbf{Example for knowing the entity:} \\
\{"5W1H": [ \\
    \{"What": "Identify the directors of 'La La Land' and 'Barbie'.", \\
    "statement": "Damien Chazelle is the director of 'La La Land', and Greta Gerwig is the director of 'Barbie'."\}, \\
    \{"Where": "Determine the nationality of Damien Chazelle and Greta Gerwig.", \\
    "statement": "Damien Chazelle is from the USA and Greta Gerwig is from the USA."\} \\
], \\
"main-point": "Both Damien Chazelle, the director of 'La La Land', and Greta Gerwig, the director of 'Barbie', are from the USA.", \\
"sub-points": [ \\
    "Damien Chazelle is the director of 'La La Land' and Gregory is the director of 'Barbie'.", \\
    "Greta Gerwig is from the USA and Damien Chazelle is from the USA." \\
], \\
"thinking": "By identifying the directors and confirming their nationalities, we verify that both Damien Chazelle and Greta Gerwig are from the USA." \} \\
\\
\textbf{Example for not knowing the entities:} \\
\{"5W1H": [ \\
    \{"What": "Identify the directors of 'La La Land' and 'Barbie'.", \\
    "statement": "Damien Chazelle is the director of 'La La Land', and PERSON is the director of 'Barbie'."\}, \\
    \{"Where": "Determine the nationality of Damien Chazelle and PERSON.", \\
    "statement": "Damien Chazelle is from the USA and PERSON is from the LOCATION."\} \\
], \\
"main-point": "Both Damien Chazelle, the director of 'La La Land', and PERSON, the director of 'Barbie', are from the LOCATION.", \\
"sub-points": [ \\
    "Damien Chazelle is the director of 'La La Land' and PERSON is the director of 'Barbie'.", \\
    "Greta Gerwig is from the USA and PERSON is from the LOCATION." \\
], \\
"thinking": "By identifying the directors and confirming their nationalities, we verify that both Damien Chazelle and PERSON are from the LOCATION." \} \\

\hline
\end{tabular}
\caption{Prompt Description for Pyramid Alignment}
\label{tab:PAprompt}
\end{minipage}
\end{table*}

\clearpage
\subsection{Parameter Setting}\label{sec:parameter}

We used a single machine with 48 cores and 256GB of memory to process the Freebase dump. The dump is managed using OpenLink Virtuoso, with the specific configuration parameters shown in the Table \ref{tab:parameter} below:

\begin{table}[H]
\centering
\begin{tabular}{ll}
\hline
\textbf{Parameter} & \textbf{Value} \\ \hline
NumberOfBuffers & 80,000,000 \\ 
MaxDirtyBuffers & 60,000,000 \\ 
MaxCheckpointRemap & 4,000 \\ 
MaxMemInUse & 64G \\ 
CheckpointInterval & 120 \\ 
ThreadsPerQuery & 16 \\ 
AsyncQueueMaxThreads & 20 \\ \hline
\end{tabular}
\caption{Virtuoso Configuration Parameters}
\label{tab:parameter}
\end{table}


\end{document}